%% file: neurips_2026.tex
\documentclass{article}

\PassOptionsToPackage{numbers, compress}{natbib}
\usepackage[main, final]{neurips_2026}
\usepackage{amsmath}
\usepackage{float}
\usepackage{multirow}
\usepackage{svg}
\usepackage[table]{xcolor}
\usepackage{comment}

\usepackage[utf8]{inputenc} 
\usepackage[T1]{fontenc}    
\usepackage{hyperref}       
\usepackage{url}            
\usepackage{booktabs}       
\usepackage{amsfonts}       
\usepackage{nicefrac}       
\usepackage{microtype}      
\usepackage{xcolor}         
\usepackage{graphicx}       
\usepackage{algorithm}
\usepackage{algpseudocode}
\usepackage{makecell}
\title{Phase-DGS: Phase-guided Dynamic Gaussian Splatting from Unsynchronized Multi-view Video}

\author{%
  \textbf{Hosung Jeon}$^{\dagger1}$ \;
  \textbf{Jun Young Jeong}$^{1}$ \;
  \textbf{Sangwoon Kwak}$^{\dagger1}$ \;
  \textbf{Joonsoo Kim}$^{1}$ \\
  \textbf{Sangmin Kim}$^{2}$ \;
  \textbf{Jaesik Park}$^{2}$ \;
  \textbf{Won-Sik Cheong}$^{1}$ \;
  \textbf{Hyon-Gon Choo}$^{1}$ \\[0.5em]
  $^{1}$Electronics and Telecommunications Research Institute (ETRI), Daejeon, Republic of Korea \\
  $^{2}$Seoul National University, Seoul, Republic of Korea \\[0.3em]
  \texttt{\{h.jeon, jyj0120, s.kwak, joonsoo.kim, wscheong, hgchoo\}@etri.re.kr} \\
  \texttt{\{sm.kim, jaesik.park\}@snu.ac.kr}}

\begin{document}

\maketitle
{\renewcommand{\thefootnote}{\fnsymbol{footnote}}%
 \footnotetext[2]{Co-corresponding authors.}}

\begin{abstract}
Spatiotemporal scene reconstruction with Dynamic Gaussian Splatting (DGS) fundamentally depends on perfect temporal synchronization across all cameras, an assumption rarely achievable in practice. Industry-standard synchronization requires expensive specialized hardware and complex setup, while data-driven alternatives including audio-based alignment, geometry matching, and pose tracking remain limited by restrictive environmental or scene-specific assumptions. To overcome these limitations, we propose Phase-DGS, a framework that replaces conventional timestamps with a semantic phase representing each frame's inherent state within the underlying motion cycle. Frames capturing the same motion state share the same phase regardless of which camera records them or when, naturally establishing spatiotemporal correspondence across unsynchronized cameras. We validate Phase-DGS under temporal offsets, frame drops, and camera freezes, achieving up to +8.5 dB PSNR improvement over baselines while enabling seamless integration with existing DGS backbones including RealTime-4DGS and FreeTimeGS.
\end{abstract}

\section{Introduction}
In recent years, radiance field representations such as Neural Radiance Fields (NeRF)~\cite{nerf2,nerf1,nerf3} and 3D Gaussian Splatting (3DGS)~\cite{3dgs1,3dgs3,3dgs2} have been extended to dynamic scenes, leading to Dynamic NeRF (DNeRF) and Dynamic Gaussian Splatting (DGS), which have emerged as a leading paradigm for spatiotemporal scene reconstruction. The latest of these methods~\cite{dnerf2,dnerf1,ftgs,4dgs} reconstruct complex real-world scenes with striking realism, enabling applications spanning from autonomous navigation to immersive content creation. However, such results come with a prerequisite: perfect temporal synchronization across all cameras, with each camera capturing the scene at exactly the same instant, frame-by-frame~\cite{syncnerf,synctrack,sync_learning2}. Without such synchronization, observations from different viewpoints no longer correspond to the same scene instant, making it ill-posed to learn a consistent radiance field under a shared timestamp. This spatiotemporal incoherence inevitably produces floaters, severe ghosting, and geometric distortion in rendered views. 

A straightforward approach is to enforce camera synchronization during the capture stage through standardized solutions, providing the radiance field with already-aligned frames that guarantee spatiotemporal coherence. Trigger-based methods such as Genlock~\cite{genlock3, genlock2, genlock1} synchronize all camera shutters via a master sync signal, achieving high precision. However, depending on the capture environment, trigger signals can be degraded by attenuation and noise, and maintaining high-quality synchronization requires expensive specialized hardware and complex wiring. Protocol-based methods such as the Precision Time Protocol (PTP)~\cite{PTP} and Network Time Protocol (NTP)~\cite{NTP} avoid dedicated wiring and offer synchronization at reasonable cost, but rely heavily on network bandwidth and stability. When transmission delays or packet losses occur, frames can be dropped, or cameras may suffer temporal drift in which their time axes gradually diverge from one another. 

Beyond capture-stage synchronization, recent works have explored training radiance fields with improved spatiotemporal coherence by using cues extracted from the captured data itself. Audio-based alignment methods~\cite{sync_audio5, sync_audio3, sync_audio2, sync_audio1, sync_audio4} synchronize cameras by matching audio signals across microphones. However, they rely on a carefully controlled acoustic environment, a condition that rarely holds in open-capture settings. Geometry- and feature-matching methods~\cite{sync_feature1, sync_feature3, sync_feature2b, sync_feature2c} estimate the degree of unsynchrony by finding spatiotemporal correspondences across views, but they struggle with weak scene texture or large viewpoint differences. Human- and pose-tracking approaches~\cite{sync_human3, sync_human5, sync_human1, sync_human2, sync_human4} align frames by tracking human body poses across cameras, but such reliance restricts their use to human-centric scenarios where a person remains visible across all frames. Reconstruction-coupled methods~\cite{syncnerf, synctrack, sync_learning2} jointly optimize sub-frame-level temporal misalignment together with the radiance field while most assign only a single offset per camera. Furthermore, these methods require camera pose information, which is typically obtained from synchronized data or provided as known calibration in the dataset. In unsynchronized recordings, however, temporal misalignment makes accurate camera pose estimation difficult.

\begin{figure}[!t]
    \centering
    \includegraphics[trim=5.6cm 6.5cm 5.6cm 6.5cm, clip,  width=1\linewidth]{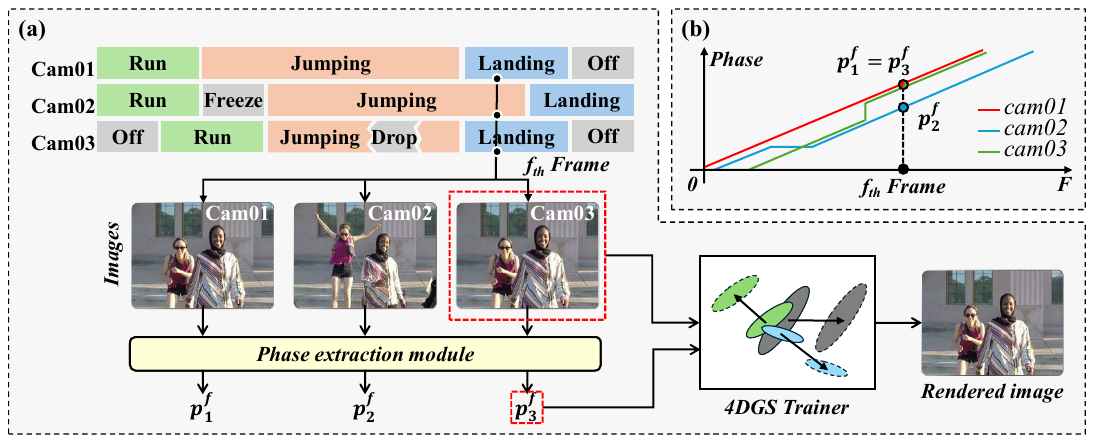}
    \caption{Conceptual diagram of Phase-DGS. (a) In an unsynchronized multi-view dataset, images sharing the frame index are actually captured at different time instants. The proposed phase extraction module estimates an inherent semantic phase from each input image, which is then used to train 4DGS. (b) The graph shows the estimated phases across cameras. When a sequence is offset, the phase shifts along the \textit{x-axis} direction; when frozen, the phase is predicted parallel to the \textit{x-axis}; and when dropped, the phase is predicted parallel to the \textit{y-axis}.}
    \label{fig:sample1}
\end{figure}

Instead of estimating absolute temporal offsets to synchronize videos, we infer an inherent semantic phase for each image that represents its position within the dynamic instance. Based on this idea, we propose Phase-DGS, a framework that integrates a phase extraction module into 4D Gaussian Splatting (4DGS) to learn a radiance field with high spatiotemporal coherence, where images sharing the same phase naturally establish spatiotemporal correspondence regardless of which camera captures them or when. As shown in Fig.~\ref{fig:sample1}, even when cameras are unsynchronized and their frame indices no longer correspond to the same instant, our phase extraction module assigns the same inherent semantic phase to images capturing the same motion state, such as the "jumping" pose, across different cameras. The estimated phases then replace absolute timestamps as input to 4DGS, allowing it to reconstruct spatiotemporally coherent rendered images. Its modular design further enables plug-and-play integration with state-of-the-art 4DGS backbones, as validated on RealTime-4DGS~\cite{4dgs} and FreeTimeGS~\cite{ftgs}. Experimentally, we validate that the proposed method optimizes 4DGS with enhanced spatiotemporal coherence under various frame-level errors, including temporal offsets, frame drops, and camera freezes. Our method achieves up to +8.5 dB improvement in PSNR over baseline methods under unsynchronized conditions.

\section{Related Works}
\textbf{Existing synchronization approaches} fall into several categories, each with its own limitations. Hardware- and software-based methods such as Genlock, Timecode, and PTP offer high precision but provide no post-hoc correction for timing errors induced by signal drift, and their cost remains a barrier to practical deployment. \textbf{Audio-based methods}~\cite{sync_audio5, sync_audio3, sync_audio1,sync_audio2,sync_audio4} align videos through shared acoustic events but rely on a controlled acoustic environment and suffer from non-negligible audio-visual latency in large-scale setups, where the finite speed of sound introduces meaningful error. \textbf{Geometry and feature matching methods}~\cite{sync_feature1, sync_feature3, visualsync, rocsync} estimate the amount of unsynchrony by finding spatiotemporal correspondences across views, and they are highly dependent on the quality of feature extraction and degrade considerably under the occlusion of tracked features. \textbf{Approaches based on human or pose tracking}~\cite{sync_human3, sync_human5, sync_human1, sync_human2, sync_human4} align frames by tracking human body poses across cameras, and strictly require the presence of humans in the scene. It substantially limits their applicability to common real-world environments. \textbf{Learning-based methods} such as Sync-NeRF~\cite{syncnerf, synctrack, sync_learning2} jointly optimize temporal offsets as learnable parameters by using rendering quality as a supervisory signal, yet they assume continuous frame availability across all views and fundamentally cannot recover from per-camera frame drops and freezes that arise independently across cameras.

\section{Method}
\subsection{Overview}
Phase-DGS replaces the absolute timestamp used in conventional 4DGS with a per-frame phase derived from the image content. We observe that each frame possesses an inherent semantic phase within video sequences, independent of its absolute timestamp. For instance, in a sit-to-stand motion, the "fully seated" pose corresponds to the same phase regardless of which camera captures it or at what moment. Based on this observation, Phase-DGS extracts a phase $p_i^f$ from each camera image $u_i^f$ and uses it instead of the timestamp $t$ to supervise 4DGS training. Note that $f$ and $i$ represent the frame and camera indices, respectively. Since the phase is derived from the image content, it naturally accommodates unsynchronized capture conditions such as frame drops, frame freezes, and temporal offsets between cameras.

\subsection{Phase Extraction Module}\label{sec:PEM}
We introduce the phase projector module, which extracts a phase $p_i^f$ from each camera image $u_i^f$ using a pretrained Vision Transformer (ViT) encoder~\cite{transformer, dinov2} followed by a lightweight MLP.

\textbf{Feature Generation.} The input image $u_i^f$ of the $f$-th frame of the $i$-th camera is coarsely cropped to a motion-intensive region of interest (ROI) and resized to match the input resolution of the ViT encoder. Then, the cropped image is passed through a pretrained ViT encoder, which partitions it into $H \times W$ patches and generates a token for each patch. We spatially aggregate all patch tokens by averaging to obtain a $d$-dimensional global feature vector $z_i^f \in \mathbb{R}^d$ that summarizes the visual content of the image:
\begin{equation}
  z_i^f = \frac{1}{HW} \sum_{n=1}^{H \times W} \left[\mathrm{ViT}_{\mathrm{enc}}\!\left(u_i^f\right)\right]_n.
\end{equation}
\textbf{Phase Projector.} The global feature vector $z_i^f$ is subsequently passed through a lightweight MLP that consists of three fully connected layers, which produces a normalized phase value $p_i^f \in [0, 1]$,
\begin{equation}
    p_i^f = \text{MLP}\left(z_i^f\right).
\end{equation}
Notably, the phase projector is trained without any explicit phase supervision; instead, it is optimized through the rendering loss together with the phase consistency and monotonic losses (see Sections~\ref{sec:loss} and~\ref{sec:training}), enabling unsupervised phase extraction directly from visual observations.

\subsection{Feature Bank} \label{sec:feature bank}
The feature bank $\mathcal{B}$ serves as a learnable reference repository that enforces cross-camera phase coherence, ensuring that identical motion states share the same phase across all views by enabling the phase projector to map each input feature to a phase through comparison against stored references. The feature bank is defined as $\mathcal{B} \in \mathbb{R}^{I \times (F \times k) \times d}$, where $I$ is the number of cameras, $F$ is the video length, $k$ is an upsampling factor, and $d$ is the dimension of the feature. $k>1$ is designed to handle videos with varying frame rates or staggered frame timestamps on some cameras, providing finer temporal resolution in the feature bank. Since the DINOv2 features are extracted independently from each image and do not contain temporal information, we make the feature bank learnable rather than fine-tuning the entire ViT encoder for every scene. Inspired by temporal prototype learning~\cite{prototype}, which aligns multiple videos through learned prototypes, the feature bank is optimized jointly with the phase projector to adapt to the scene, so that frames capturing the same motion state are assigned the same phase across all views. When the feature bank is fixed, which corresponds to a fixed-memory alternative, the accuracy of phase estimation drops substantially (see Section~\ref{sec:ablation}).

\subsection{Loss Function} \label{sec:loss}
Phase-DGS is trained with a combination of three loss functions,
\begin{equation}
    \mathcal{L} = \lambda_{phase}\mathcal{L}_{phase} + \lambda_{mono}\mathcal{L}_{mono} + \lambda_{render}\mathcal{L}_{render}.
\end{equation}

\textbf{Phase Consistency Loss.} $\mathcal{L}_{phase}$ supervises the phase projector so that it retrieves features with matching phases from the feature bank. Given an input image $u_i^f$, the extracted feature $z_i^f$ is passed through the phase projector to obtain the phase $p_i^f$. The entire $i$-th camera array of the feature bank, $\mathcal{B}_i \in \mathbb{R}^{(F \times k) \times d}$, is then passed through the phase projector to generate a phase map $\mathcal{P}_i \in \mathbb{R}^{(F \times k) \times 1}$. An attention weight $w$ is computed by applying softmax to the distance between the phase map and $p_i^f$,
\begin{equation}
    w = \text{softmax}\left(-\left|\mathcal{P}_i - p_i^f\right|\right).
\end{equation}
The reference feature $\hat{z}_i^f$ that shares the same phase as $p_i^f$ is then retrieved through a weighted sum over the feature bank,
\begin{equation}
    \hat{z}_i^f = \sum w \cdot \mathcal{B}_i.
\end{equation}
The loss is defined as the MSE between $z_i^f$ and $\hat{z}_i^f$,
\begin{equation}
    \mathcal{L}_{phase} = \text{MSE}\left(z_i^f, \hat{z}_i^f\right).
\end{equation}

\textbf{Monotonic Loss.} $\mathcal{L}_{mono}$ promotes the phase map generated from the feature bank to increase monotonically along the temporal axis. The entire feature bank is passed through the phase projector to produce $\mathcal{P} \in \mathbb{R}^{I \times (F \times k) \times 1}$, and the difference between adjacent phases is computed for each camera,
\begin{equation}
    \Delta_i^{f} = \mathcal{P}_i^{f+1} - \mathcal{P}_i^{f}, 
    \quad f = 0, 1, \ldots, Fk - 2.
\end{equation}
A penalty is applied whenever the phase difference falls below the minimum increment $a = \frac{1}{F \times k}$,
\begin{equation}
    \mathcal{L}_{mono} = \sum_{i}\sum_{f} 
    \text{relu}\left(-\Delta_i^{f} + \frac{1}{Fk}\right).
\end{equation}
This term encourages the phase projector to learn temporally consistent phases across the feature bank. 

\textbf{Rendering Loss.} $\mathcal{L}_{render}$ optimizes the rendering quality of 4DGS by minimizing the visual discrepancy between the rendered images and the ground-truth observations.

\begin{figure}[H]
    \centering
    \includegraphics[trim=0.0cm 2.8cm 0.0cm 2.8cm, clip, width=1\linewidth]{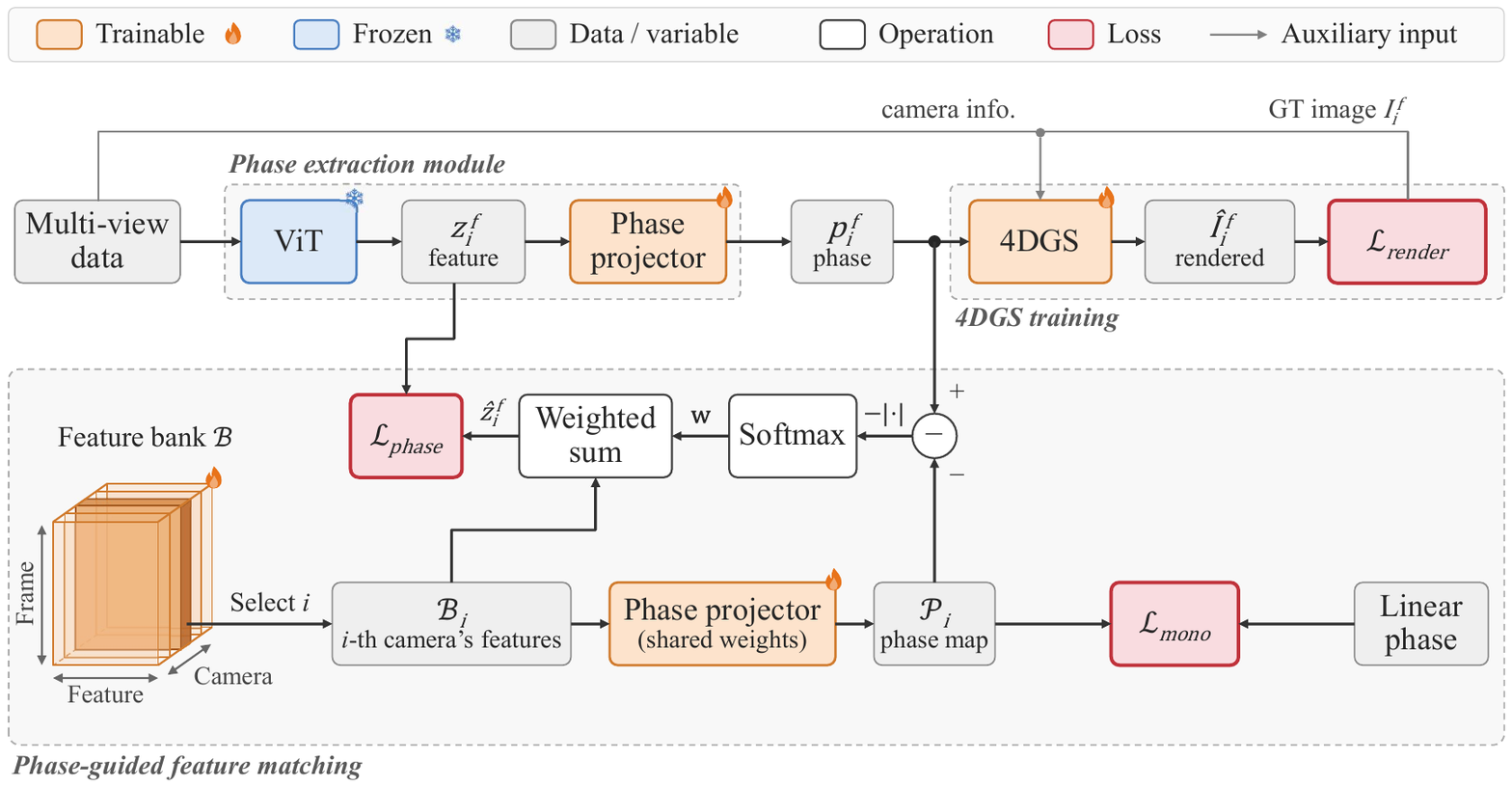}
    \caption{Training pipeline of Phase-DGS. Multi-view data provides camera parameters (view matrix, focal length, resolution) and an RGB image. The phase projector module predicts a phase value from the RGB image, replacing the timestamp input of the 4DGS backbone. The feature bank serves as a learnable reference repository that enforces cross-camera phase coherence, so that identical motion states share the same phase across all views. Training is guided by three losses: a Rendering loss between the rendered and input images, a monotonic loss, and a phase consistency loss.}
    \label{fig:sample2}
\end{figure}

\subsection{Integration with Dynamic Gaussian Splatting} \label{sec:integration}
Existing 4DGS methods model the deformation of 3D Gaussians over time by taking the absolute timestamp $t$ of each frame as input~\cite{4dgs, ftgs}. However, as discussed above, absolute timestamps are inherently unreliable in real-world capture environments because of inter-camera synchronization errors, frame drops, and frame freezes.

To address this issue, we replace the timestamp $t$ in 4DGS model $\mathcal{G}$ with the phase $p_i^f$ extracted by Phase-DGS, as shown in Fig.~\ref{fig:sample1}. Since $p_i^f$ is extracted independently from the image content of each camera, it naturally compensates for inter-camera synchronization errors, frame drops, and frame freezes without requiring any explicit timestamp correction. In other words, whereas conventional 4DGS optimizes Gaussians along the temporal axis, Phase-DGS optimizes Gaussians along the phase axis,
\begin{equation}
    \mathcal{G}(p_i^f) \leftarrow \mathcal{G}(t).
\end{equation}
Note that the phase is designed as a one-dimensional scalar because it plays the same role as the timestamp $t$. Since existing 4DGS methods take a scalar timestamp as input~\cite{4dgs, ftgs}, the phase can be integrated in a plug-and-play manner without modifying the backbone. Furthermore, all objects within a frame are captured at the same instant, so a single phase per frame is sufficient to represent the temporal state of the scene regardless of the number of moving objects. The motion of each object, including complex motions and multi-object interactions, is still modeled by the deformation of the 4DGS backbone, as in the timestamp-based setting.

\subsection{Training}\label{sec:training}
Phase-DGS is trained end-to-end by jointly optimizing a plug-and-play 4DGS backbone as shown in Fig. ~~\ref{fig:sample2} (e.g., RealTime-4DGS~\cite{4dgs}, FreeTimeGS~\cite{ftgs}), the phase projector, and the feature bank under the combined loss in Eq.~(3). The ViT encoder remains frozen throughout the training, while all other modules are updated by gradient descent.
In each iteration, a frame $u_i^f$ is sampled from the camera $i$ and passed through the ViT encoder and the phase projector to produce the phase $p_i^f$. The estimated phase replaces the absolute timestamp $t$ as the temporal coordinate for the 4DGS backbone, and all three losses are back-propagated simultaneously.

\section{Experimental results}

\subsection{Implementation Details.}

\textbf{Phase Extraction Module.}
For the ViT encoder, we adopt DINOv2~\cite{dinov2} with the \texttt{dinov2\_vitb14} configuration. The phase projector~(see Section \ref{sec:PEM}) MLP follows a $d \rightarrow 256 \rightarrow 128 \rightarrow 64 \rightarrow 1$ architecture, in which each hidden layer is activated by GELU and the final output is normalized to $[0, 1]$ by a Sigmoid activation. 

\textbf{4DGS Backbones.}
We primarily adopt FreeTimeGS~\cite{ftgs}, which we re-implement based on the original paper, and we additionally validate plug-and-play compatibility with RealTime-4DGS~\cite{4dgs}. These two methods constitute the state of the art in 4DGS and represent the \textit{explicit} and \textit{implicit} paradigms of modeling Gaussian motion, respectively. 

\textbf{Training Details.} Phase-DGS is trained with the Adam optimizer and a batch size of one camera per iteration. The shared hyperparameters across both backbones are a learning rate of $1 \times 10^{-4}$ for the phase projector, a learning rate of $1 \times 10^{-3}$ for the feature bank, $\lambda_{render} = 1.0$, and $\lambda_{phase} = 0.1$. The backbone-specific configurations are as follows. FreeTimeGS is trained for 500,000 iterations with an upsampling factor $k = 2$ and $\lambda_{mono} = 0.1$, while RealTime-4DGS is trained for 60,000 iterations with $k = 3$ and $\lambda_{mono} = 0.01$. Compared with the FreeTimeGS backbone alone, Phase-DGS requires approximately 30\% more GPU memory and $2.1\times$ longer training time per iteration, while the rendering cost at inference remains the same (see Section~A).

\textbf{Datasets.}
We evaluated Phase-DGS on two representative datasets for dynamic scene reconstruction, each featuring distinct camera configurations to ensure a comprehensive assessment.
The Nvidia Dynamic Scene Dataset~\cite{nvidia_dataset} consists of 12 front-facing Full HD cameras, and SelfCap~\cite{selfcap_dataset} provides 24 convergent 4K cameras, with sequences of approximately 100 and 300 frames, respectively. For the Nvidia Dynamic Scene Dataset, we use three scenes: Jumping, Playground, and Umbrella. For SelfCap, we use Corgi, Yoga, and Bike.
For both datasets, all images are downsampled to $960 \times 540$. To validate robustness under various unsynchronized conditions, we evaluated on three data configurations. The \textbf{Synced} condition uses the provided dataset as is and assumes synchronized cameras. The \textbf{Shifted} condition introduces inter-camera temporal offsets of about 20--30 frames by randomly shifting the starting frame of each camera. The \textbf{Dropped} condition simulates severe real-world asynchrony with 10\% random frame drops and 5--10 frames freeze, which vary independently across cameras.

\textbf{Train and Test View Selection.} Since Phase-DGS improves the optimization quality of the radiance field, we select a single camera located at the outermost spatial position in each scene as the test view. This view has the largest viewpoint difference from the train views, where artifacts such as spatial incoherence caused by inaccurate timestamps appear more prominently.

\subsection{Baselines}
We compare Phase-DGS to the following baselines. \textbf{4DGS w/o Sync} trains the 4DGS backbone directly using absolute timestamps without any synchronization and serves as a lower-bound reference. \textbf{VisualSync}~\cite{visualsync} performs video synchronization as a preprocessing step, after which the 4DGS backbone is trained on the aligned sequences. \textbf{Sync-NeRF}~\cite{syncnerf} jointly optimizes temporal offsets together with neural radiance field reconstruction in an end-to-end manner.

\subsection{Quantitative Results}
As shown in Tab.~\ref{Table_Avg}, the Synced condition achieves the highest performance in all datasets (all experimental results are shown in Tab.~\ref{Table_SelfCap} and \ref{Table_Nvidia}, and see evaluation detail in Section~\ref{Evaluation Protocol}). In contrast, the Shifted and Dropped conditions exhibit different tendencies depending on the dataset. A noteworthy observation is that RealTime-4DGS~\cite{4dgs}, which implicitly represents motion, is relatively more robust to unsynchronized conditions than FreeTimeGS~\cite{ftgs}, which explicitly represents motion. Due to this property, the proposed Phase-DGS shows only a marginal improvement in PSNR when combined with the RealTime-4DGS backbone. In contrast, when combined with the FreeTimeGS backbone, which uses an explicit motion representation, Phase-DGS achieves a substantial improvement in quantitative metrics. FreeTimeGS is therefore particularly sensitive to timestamp inaccuracy, and under such conditions the phase-based supervision of Phase-DGS achieves the greatest effect. 

Since Sync-NeRF~\cite{syncnerf} is retrained under capture conditions that differ from those in the original paper, both the COLMAP-based pose initialization and hyperparameter settings additionally affect the reconstruction quality under all conditions. In particular, the SelfCap dataset~\cite{selfcap_dataset} proves incompatible with the default configuration of Sync-NeRF, which was originally designed for forward-facing captures such as the Nvidia Dynamic Scene dataset~\cite{nvidia_dataset}, resulting in notably degraded reconstruction quality. Furthermore, the Dropped condition falls outside the design scope of Sync-NeRF, which assumes only real-valued sub-frame temporal offsets. Consequently, Sync-NeRF also yields poor performance under the Dropped condition.

VisualSync~\cite{visualsync} successfully corrects synchronization errors before training in the Shifted conditions, yielding reconstruction quality comparable to that of the backbone under the Synced condition. However, the Dropped condition disrupts the temporal consistency of feature correspondences, preventing VisualSync from reliably estimating epipolar geometry. In particular, even under the Synced condition, VisualSync introduces an unexpected frame offset of approximately 3\% of the total frame count after synchronization, which decreases the reconstruction quality.

Under the Synced condition, Phase-DGS improves the average PSNR on SelfCap by 0.55\,dB and 1.29\,dB with the FreeTimeGS and RealTime-4DGS backbones, respectively, while it shows a slightly lower PSNR on the Nvidia Dynamic Scene dataset. In SelfCap, the convergent multi-camera setup introduces greater viewpoint diversity, making spatiotemporal coherence a more critical factor for reconstruction quality. As a result, accurate synchronization leads to a more pronounced improvement in rendering quality. In contrast, in the Nvidia Dynamic Scene dataset, the simpler forward-facing configuration allows the backbone to converge quickly in terms of image quality, whereas Phase-DGS requires additional iterations for phase learning within the same training budget, resulting in a slightly lower PSNR than the backbone under the Synced condition. We note that Phase-DGS is designed to provide robustness to synchronization errors rather than to improve reconstruction under perfect synchronization. Since the synchronization status of real-world captures is often unknown, it is noteworthy that Phase-DGS remains within 0.8\,dB of the backbone under the Synced condition, while improving the PSNR by up to 6.5\,dB under the Dropped condition (SelfCap, FreeTimeGS).

\input{Tables/Table_finalfinal}

\begin{figure}[!t]
    \centering
    \includegraphics[trim=7.2cm 4.3cm 7.9cm 4.6cm, clip, width=1\textwidth]{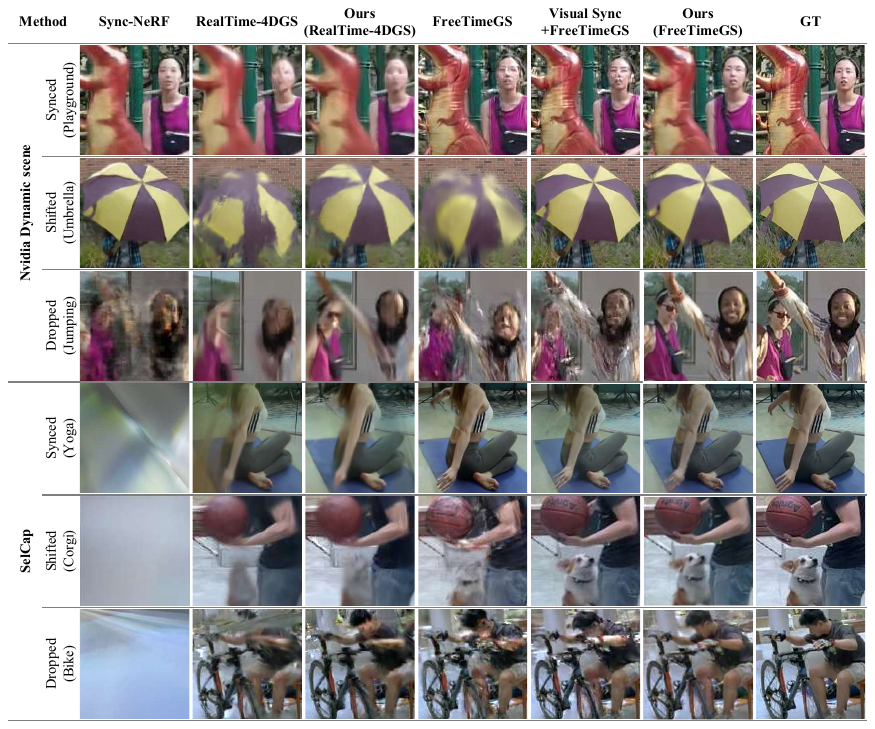}
    \caption{Qualitative comparison under the synchronized condition (Synced) and the unsynchronized condition (Shifted and Dropped) 
    on \textbf{Nvidia Dynamic Scene dataset} and \textbf{SelfCap datasets}. Phase-DGS consistently produces cleaner renderings while baselines exhibit ghosting and 
    geometric distortion.}
    \label{fig:Visualization1}

    \centering
    \includegraphics[trim=0cm 3.5cm 0cm 2.0cm, clip, width=0.95\textwidth]{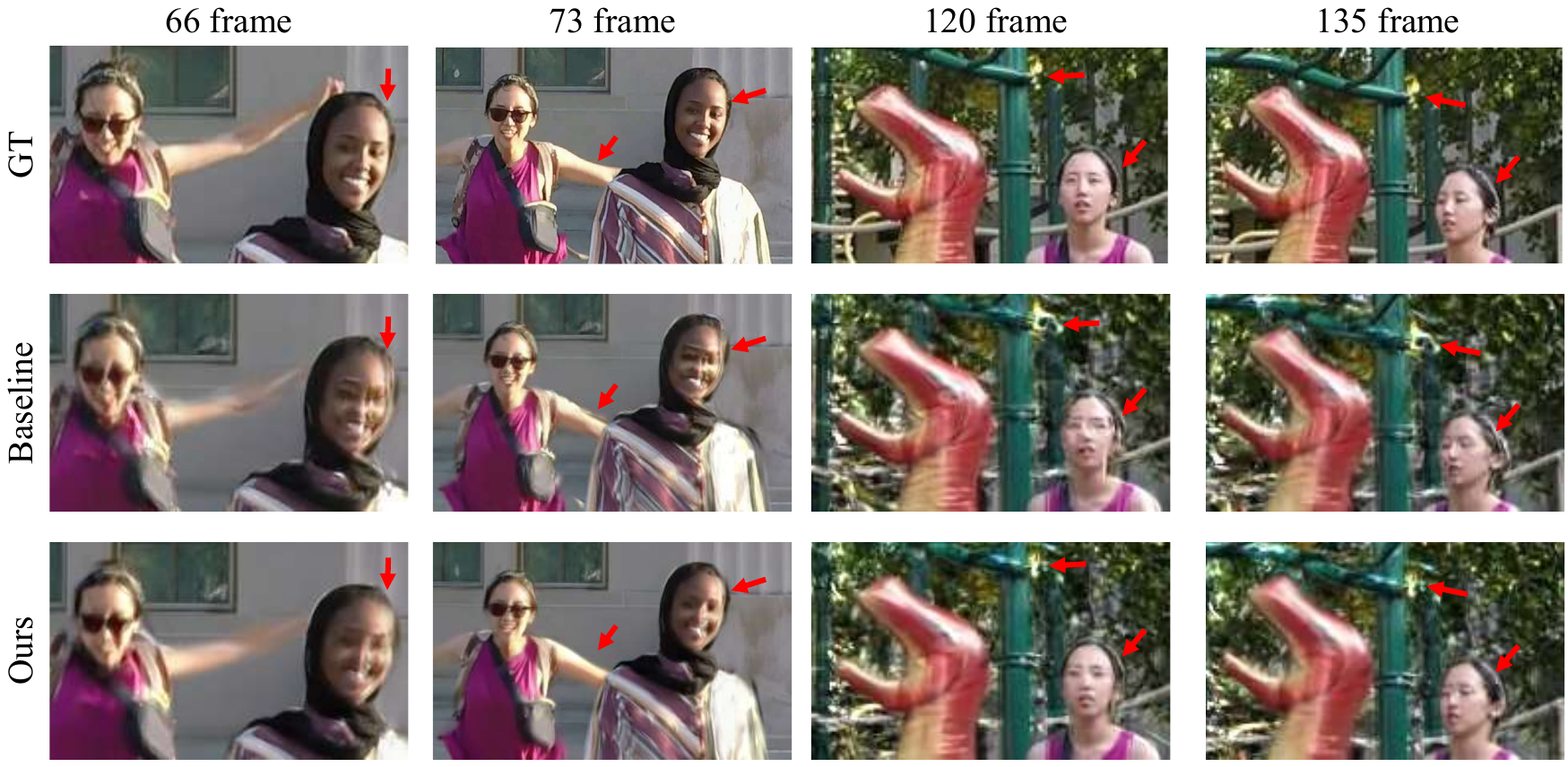}
    \caption{Qualitative comparison with enlarged views under the Synced 
condition. While baseline models exhibit ghosting and geometric 
distortion, Phase-DGS consistently reduces artifacts in rendered images. 
\textcolor{red}{$\rightarrow$}: visual artifact.}
    \label{fig:Qualitative}
\end{figure}

\subsection{Qualitative Results}
As shown in Fig.~\ref{fig:Visualization1}, Phase-DGS with the FreeTimeGS~\cite{ftgs} backbone achieves the best visual quality among all the methods compared. This qualitative comparison reveals perceptual differences that quantitative metrics alone cannot fully capture. For the RealTime-4DGS~\cite{4dgs} backbone, as an additional discussion~, the difference in PSNR before and after applying Phase-DGS is small. However, the rendering results show that the Gaussians in dynamic regions are accurately synchronized and converge to a single spatial point, and as a result the shape of dynamic objects appears much sharper than in the baseline. Since the phase estimated by the phase projector is gradually reflected throughout the training process, the overall PSNR of the scene shows only a limited change, while the Gaussian alignment in dynamic regions steadily improves. These results demonstrate that Phase-DGS effectively learns accurate phase information for each frame, precisely defining the three-dimensional position of each Gaussian and enabling cleaner renderings across both backbones.

Fig.~\ref{fig:Qualitative} presents enlarged views for a more detailed comparison, where all outputs are rendered from models trained under the Synced condition. Comparing Phase-DGS against the FreeTimeGS backbone, we show pairs of frames from the same scene to highlight inter-frame differences. Although Phase-DGS under the Synced condition yields a lower PSNR than the FreeTimeGS baseline in the Jumping scene, the rendered images reveal noticeably fewer visual artifacts, a trend that is consistently observed in the Playground scene as well. We attribute this to the fact that Phase-DGS learns a radiance field with higher spatiotemporal coherence, enabling more geometrically consistent reconstructions even when quantitative metrics do not fully reflect this improvement. 
A notable observation is that even on datasets provided as synchronized, the backbone models without Phase-DGS still exhibit subtle rendering artifacts, empirically supporting the claim that even sub-frame-level timing disparities can measurably affect reconstruction quality.

\subsection{Ablation Study}\label{sec:ablation}
We conduct ablation studies on the Jumping scene of the Nvidia Dynamic Scene Dataset~\cite{nvidia_dataset} and the Bike scene of SelfCap~\cite{selfcap_dataset} under the Dropped condition to analyze the contribution of each component in Phase-DGS with FreeTimeGS~\cite{ftgs} backbone.

\textbf{Monotonic Loss.} Without this loss, the phase is estimated in a temporally disordered manner, causing the beginning and end of the sequence to be assigned phases without consistent ordering.  (see Section \ref{sec:loss}) 

\textbf{Feature~ Bank.} When the feature bank is initialized but not updated during training, the stored features remain fixed and the accuracy of phase estimation drops substantially (see Section \ref{sec:feature bank}). This shows that the learnable feature bank plays an essential role by providing a sufficient reference over the entire sequence of each camera, which stabilizes the phase estimation process. 

\textbf{Upsampling Factor $k$.} We vary the upsampling factor $k$ of the feature bank and analyze its effect on performance (see Section \ref{sec:feature bank}). When $k=1$, the feature bank operates at the original frame rate and cannot capture sub-frame asynchrony. At $k=2$, the finer temporal resolution yields the best trade-off between phase accuracy and computational cost. Increasing $k$ to 4 does not improve performance, as the enlarged feature bank requires more training iterations to converge within the fixed training budget, leading to a slight PSNR degradation. 

\textbf{ViT Backbone.} We compare DINOv2-S, -B, and -L~\cite{dinov2} depending on the number of parameters of the ViT encoder. The self-supervised DINOv2 learns detailed visual representations, and the performance gap between different sizes of DINOv2 models is small. This suggests that phase estimation captures the overall progression of motion, for which the global semantic representation learned by the base model is already sufficient. 

\textbf{ROI Crop.} We compare two input configurations. The first crops the input image to the motion-intensive ROI, while the second forcibly resizes the full image to $224 \times 224$ for the visual transformer (see Sections~\ref{sec:PEM}  and~\ref{sec:training})~\cite{dinov2}. The ROI-based configuration achieves slightly better performance, but the difference is less than 0.2\,dB in both scenes, which shows that Phase-DGS also operates on full images without the ROI crop. The ROI crop is expected to be more effective when the dynamic objects occupy only a small region of the image, since resizing the full image to $224 \times 224$ can remove fine motion cues.

\input{Table_Ablation}

\section{Conclusion}\label{sec:conclusion}
We presented Phase-DGS, a self-supervised framework for robust dynamic scene reconstruction under unsynchronized multi-camera conditions. By replacing absolute timestamps with per-frame phases extracted directly from images, Phase-DGS eliminates the reliance on hardware synchronization and enables robust 4DGS training even under severe frame drops and temporal offsets. Experiments on Nvidia Dynamic Scene Dataset and SelfCap demonstrate that Phase-DGS consistently outperforms existing synchronization baselines under unsynchronized conditions, achieving competitive performance even under the synchronized setting. Furthermore, plug-and-play compatibility with multiple 4DGS backbones highlights the generalizability of our approach.  The ability to train without precise timestamps suggests the potential to extend Phase-DGS to archival or previously captured footage, where hardware synchronization cannot be applied retroactively. We hope Phase-DGS serves as a practical and scalable solution for real-world multi-camera capture systems.

\textbf{Limitations and Future Work.} The reconstruction quality of Phase-DGS depends on the underlying 4DGS backbone, and the way timestamps are processed inside the network also affects performance. In addition, Phase-DGS currently struggles with scenes containing periodic motions, such as the \textit{Balloon1} sequence, where visually similar frames make accurate phase assignment difficult. We plan to address these issues through backbone-agnostic training strategies and inter-frame constraints that encourage smooth and consistent phase transitions. 

\section*{Acknowledgements}
This work was supported by the IITP grant funded by the Korea government (MSIT): Development of immersive video spatial computing technology for ultra-realistic metaverse services (No.2022-0-00022, RS-2022-II220022), and by the IITP grant (RS-2021-II211343, AI Graduate School Program (Seoul National University)), funded by the Korean government (MSIT).

\section*{Disclosures}
The authors declare no competing interests.

\bibliographystyle{plainnat}
\bibliography{references}

\appendix
\setcounter{figure}{0}
\setcounter{table}{0}
\renewcommand{\thefigure}{A\arabic{figure}}
\renewcommand{\thetable}{A\arabic{table}}
\renewcommand{\theHfigure}{A\arabic{figure}}  
\renewcommand{\theHtable}{A\arabic{table}}    

\newpage

\section{Additional Implementation Details}\label{sec:impl}

\begin{figure}[!t]
    \centering
    \includegraphics[trim=4cm 6cm 4cm 6cm, clip, width=1.0\linewidth]{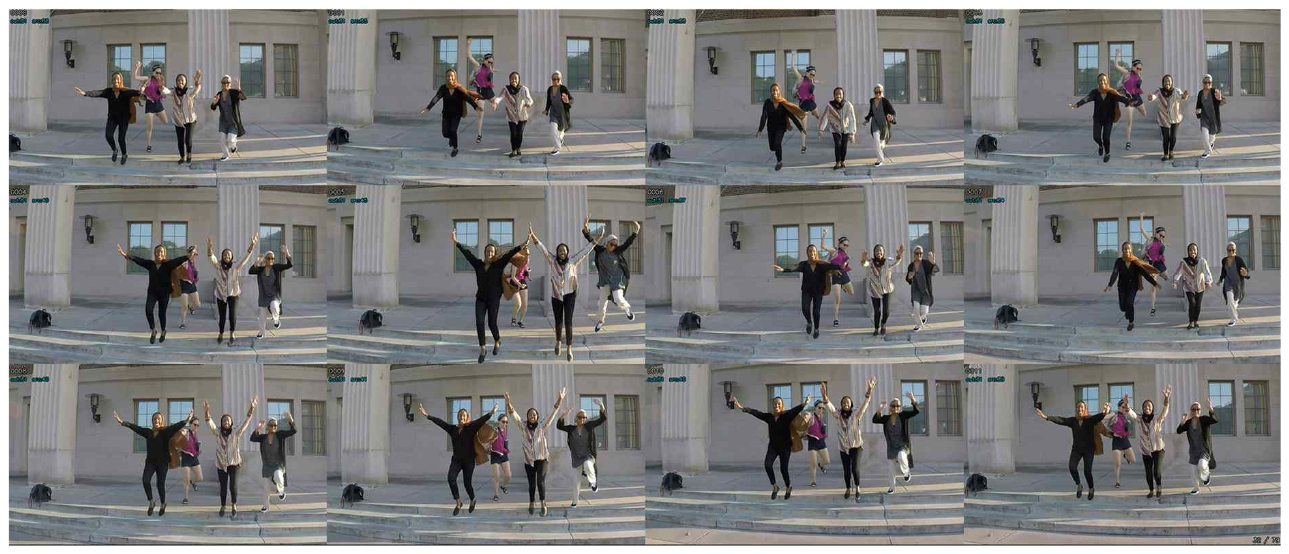}
    \caption{Multi-camera frames at the same frame index from the \textit{Jumping} sequence (Nvidia Dynamic Scene Dataset) under the \textbf{Dropped} condition. Despite sharing the same frame index, each camera captures a distinct motion state, illustrating the severity of temporal 
  misalignment introduced by frame drops and freezes.}
    \label{fig:videos}
\end{figure}

\textbf{Data Generation.}
To assess robustness under various real-world capture conditions, we construct three data configurations with increasing levels of temporal asynchrony.
The \textbf{Synced} condition uses the original dataset without modification, serving as the upper-bound reference under ideal synchronization.
The \textbf{Shifted} condition introduces inter-camera temporal offsets of 20--30 frames by randomly shifting the starting frame of each camera, simulating clock drift or delayed recording onset.
The \textbf{Dropped} condition simulates severe asynchrony by applying 10\% random frame drops and 5--10 frame freezes that vary independently across cameras, reflecting realistic packet loss and buffer failures in large-scale capture systems.
As illustrated in Fig.~\ref{fig:videos}, frames captured at the same frame index across cameras under the \textbf{Dropped} condition exhibit pronounced motion discrepancies, highlighting the severity of temporal misalignment introduced by our simulation.
Additional visualizations are provided in the supplementary video.

\begin{figure}[!t]
    \centering
    \includegraphics[trim=0cm 4.0cm 0cm 4.0cm, clip, width=0.7\linewidth]{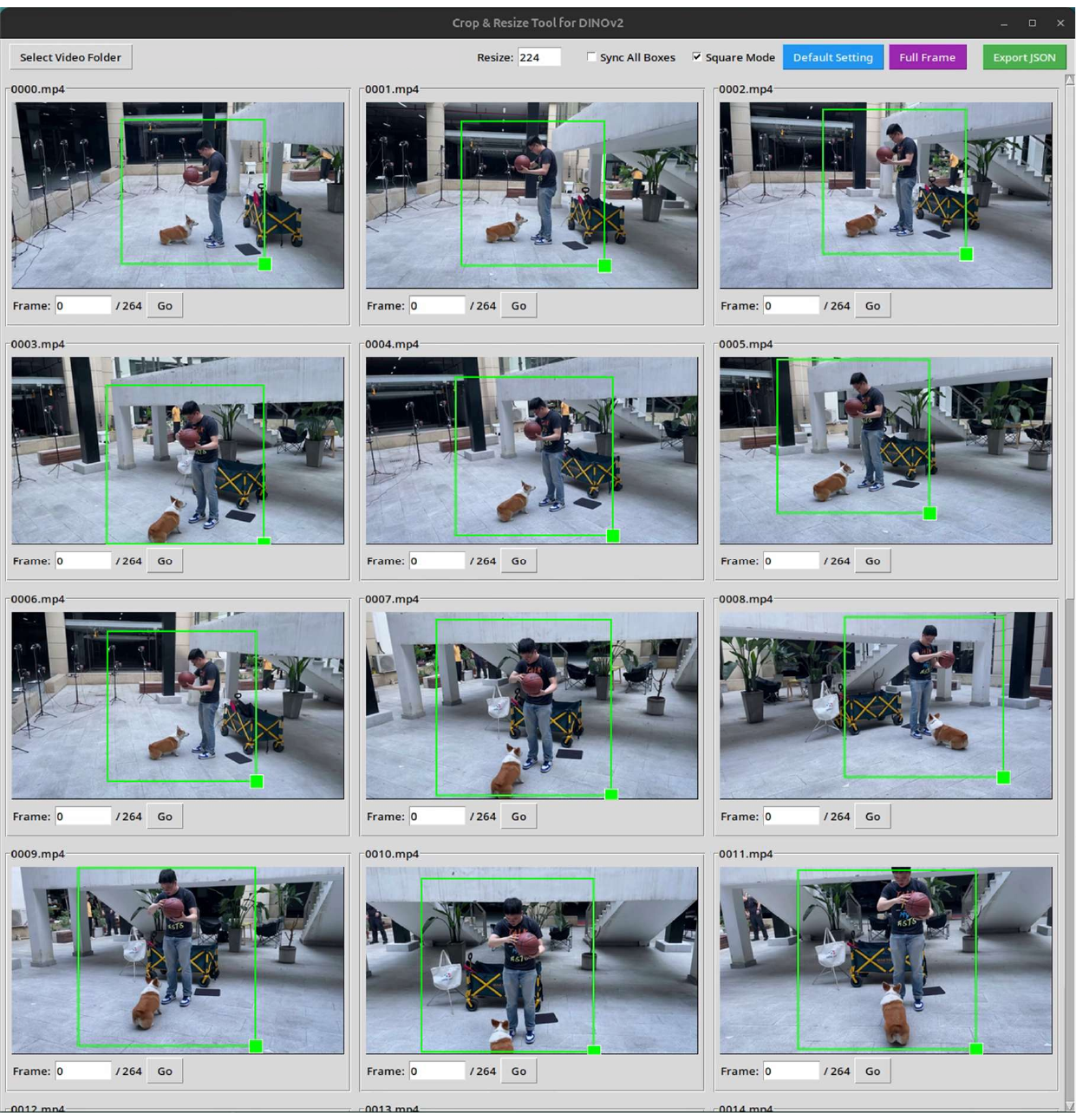}
    \caption{Interface of the ROI extraction tool used in our experiments. The tool supports manual bounding box selection, with options to use the central square region or the full frame. Different cameras in the same scene can have independent bounding boxes.}
    \label{fig:roi_tool}
\end{figure}

\textbf{ROI Cropping.} Since this work focuses on the synchronization problem, we implement a manual ROI extraction tool throughout the experiments with the help of an LLM. The ROI extraction tool used in our experiments is implemented as a simple interface that supports manual bounding box selection as shown in Fig.~\ref{fig:roi_tool}. For each scene, the ROI is specified to cover the region where motion is most concentrated, based on a single representative frame from the scene. The tool also provides options to use the central square region of the frame or to keep the full frame without cropping. Different cameras in the same scene can have bounding boxes of different sizes and positions, and the location of each bounding box can be specified freely. Note that the ROI crop is an optional preprocessing step and is not required by Phase-DGS (see Section~4.5). The ROI extraction process can be further automated using algorithms such as optical flow, motion detection, or video segmentation models such as SAM2~\cite{sam2}. For reproducibility, we will release the ROI coordinates and segmentation masks used in our experiments together with the code.

\textbf{Feature bank Initialization.} The feature bank is initialized with features extracted from all frames of all cameras. When the upsampling factor is $k > 1$, the feature entries at upsampled positions that do not correspond to original frames are filled by linear interpolation between neighboring features. Initializing the feature bank in this way provides meaningful feature values from the beginning of training, which shortens the convergence time.

\textbf{Compute Resources.} All experiments in this work were conducted on a single NVIDIA RTX 5090 GPU with Python 3.12, PyTorch 2.8.0, and CUDA 12.8. For the RealTime-4DGS~\cite{4dgs} backbone, 100K iterations take approximately one hour. For the FreeTimeGS~\cite{ftgs} backbone, 500K iterations take approximately 2.5 hours. The GPU memory usage depends on the scene, with the RealTime-4DGS backbone using approximately 12.4 GB and the FreeTimeGS backbone using approximately 4 GB. The full set of reported experiments covers 6 scenes (3 from the Nvidia Dynamic Scene Dataset and 3 from SelfCap) across 3 synchronization conditions (Synced, Shifted, Dropped) and two 4DGS backbones (RealTime-4DGS and FreeTimeGS). The reported computation does not include preliminary experiments and failed runs.

\textbf{Training Overhead.}
To quantify the additional cost of phase learning, we measure the GPU memory usage and training throughput with the FreeTimeGS~\cite{ftgs} backbone on a multi-view sequence captured with 23 cameras (50 frames), as shown in Tab.~\ref{tab:overhead}. Note that the absolute values differ from those reported above because the dataset is different. Phase-DGS requires an additional 514\,MB (+30.5\%) of GPU memory, which mainly comes from the feature bank, the phase projector, and the ViT encoder. Since the ViT encoder remains frozen, back-propagation through the encoder is not required, and only the lightweight phase projector and the feature bank are additionally optimized. The training throughput decreases from 148 to 70 iterations per second ($2.11\times$ slower). This overhead applies only to training, since the ViT encoder, the phase projector, and the feature bank are not used at inference, where the backbone is rendered directly at a given phase.

\begin{table}[h]
  \centering
   \caption{Training overhead of Phase-DGS with the FreeTimeGS backbone on a 23-camera, 50-frame sequence, measured on a single NVIDIA RTX 5090 GPU.}
  \label{tab:overhead}
  \begin{tabular}{lcc}
    \toprule
    Configuration & GPU Memory & Training Throughput \\
    \midrule
    FreeTimeGS (backbone only) & 1686\,MB & 148\,it/s \\
    Phase-DGS (FreeTimeGS)     & 2200\,MB (+514\,MB) & 70\,it/s ($2.11\times$ slower) \\
    \bottomrule
  \end{tabular}
\end{table}

\newpage

\section{Evaluation Protocol} \label{Evaluation Protocol}
\begin{figure}[!t]
    \centering
    \includegraphics[trim=5.8cm 4.7cm 5.0cm 4.0cm, clip, width=1\linewidth]{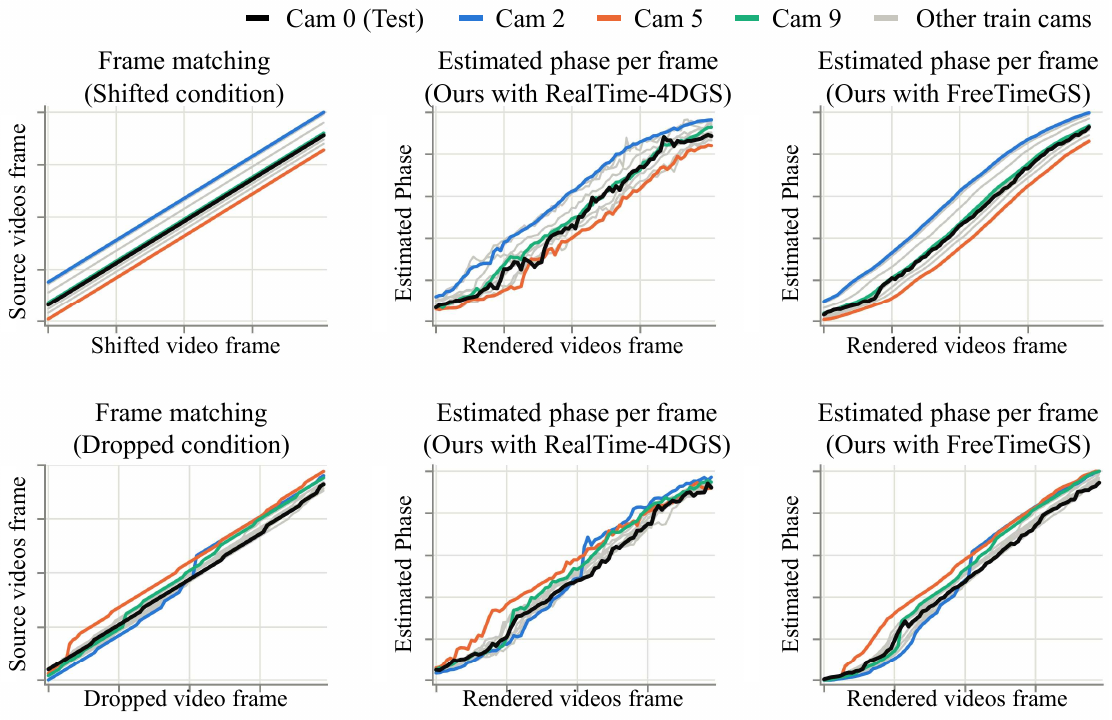}
    \caption{Phase estimation results on the Jumping scene under the Shifted (top) and Dropped (bottom) conditions. (Left) Ground-truth frame correspondences between each camera and the source video. (Center, Right) Per-frame phase estimates produced by Phase-DGS with the RealTime-4DGS and FreeTimeGS backbones, respectively.}
    \label{fig:phase_graph}
\end{figure}

\textbf{Phase Reference Computation.} Phase-DGS does not estimate an absolute timestamp from an image but instead extracts a relative phase value. As shown in Fig.~\ref{fig:phase_graph}, the phase values produced by Phase-DGS for frames within a single scene follow a non-linear curve. It is noted that the phase curves of the train views are regularized by the monotonic loss so that they remain monotonic, while the test view (Cam 0) is not subject to this loss and thus exhibits a slightly different phase curve, although it follows a similar monotonic trend. As a result, the phase stamp of the Phase-DGS, which refers to the ordering of frames along the phase axis and plays a role similar to a time stamp, does not provide absolute temporal information. To quantitatively evaluate Phase-DGS, we therefore need to define a reference phase stamp for each scene.
We should construct the reference phase stamp for the test view. The unsynchronized dataset used in our experiments provides the correspondence between each view's frame index and the index in the original source video. Using this correspondence, we can identify which frame of the source video each test view frame is sampled from, and then find the train view frames that are sampled from the same source index. The phase reference for each test view frame is obtained by averaging the phases of the corresponding train view frames. The overall procedure is summarized in Algorithm~\ref{alg:phase_eval}.
Since this protocol does not recover the exact absolute phase stamp but only estimates a reference phase in an indirect manner, the actual time of a test view frame may not perfectly align with the time implied by the estimated reference phase. This misalignment can introduce minor inaccuracies in the reported quantitative metrics. We note that this protocol is used only for evaluation purposes, and the exact procedure for constructing the phase-line may vary depending on the application or the dataset. However, this correspondence is a form of ground-truth temporal alignment that is available only in our simulated datasets, and it is not used for the baselines. The reported metrics therefore reflect the reconstruction quality of Phase-DGS given an accurate temporal anchor, which is stronger than the anchor available in practice, where the global phase stamp is defined without ground truth (see Section~\ref{sec:discussion}).

\begin{figure}[!t]
    \centering
    \includegraphics[trim=0cm 0.1cm 0cm 0.1cm, clip, width=1.0\linewidth]{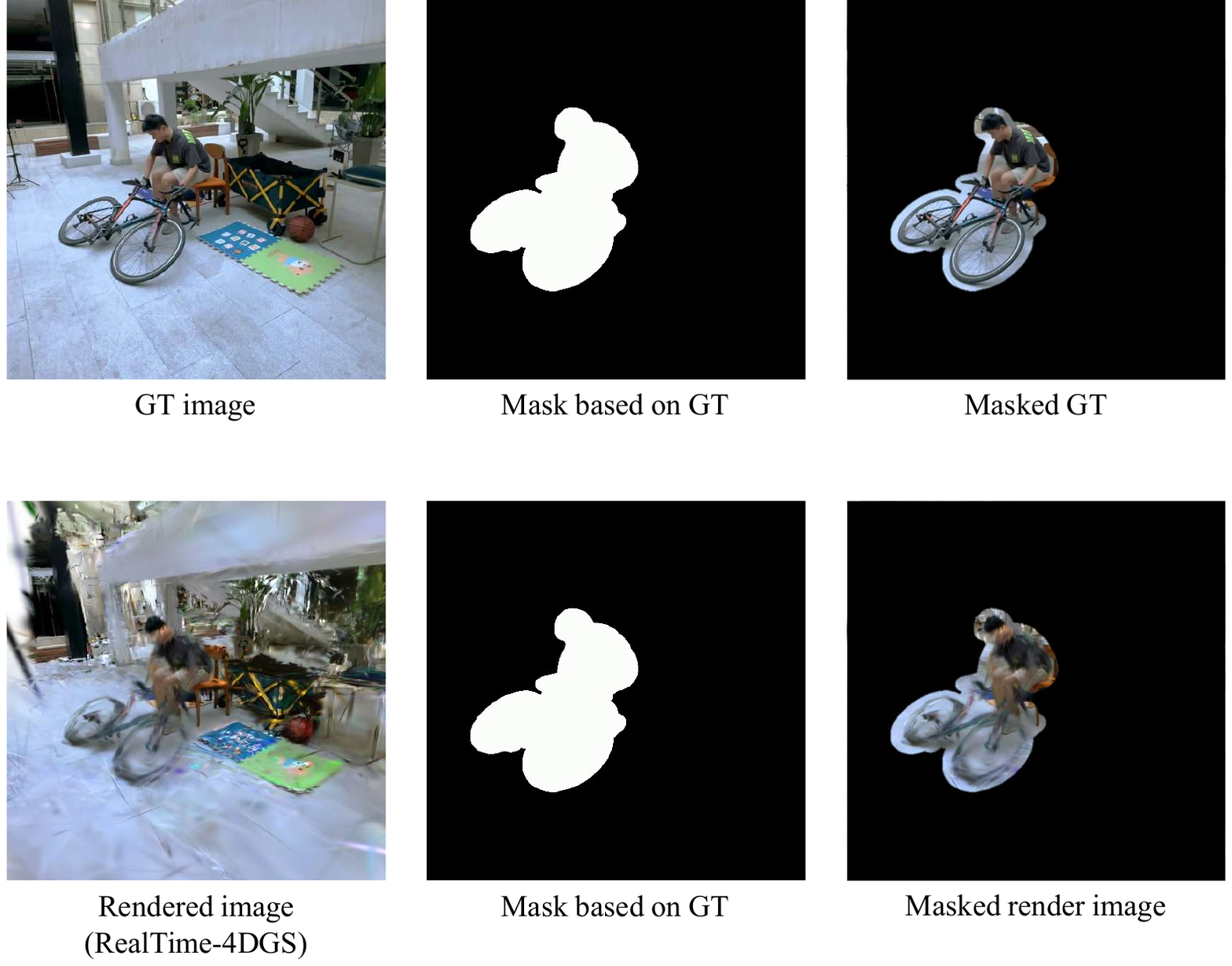}
    \caption{To assess rendering quality in dynamic regions, we generate segmentation masks from ground-truth images and compute metrics exclusively over the masked areas. The mask is generated by object segmentation using SAM2~\cite{sam2}, and a small padding margin is applied to the mask boundaries to assess the reconstruction quality of regions surrounding dynamic objects.}
    \label{fig:mask}
\end{figure}

\textbf{Evaluation.} To evaluate the performance of the trained GS model, we first compute the phase reference $\tilde{p}_{test}^f$ for each test view frame using Algorithm~\ref{alg:phase_eval}. Here, the reference phase $\tilde{p}_{test}^f$ from Algorithm~\ref{alg:phase_eval} contains estimation noise, which can lower the reported metrics even when the underlying reconstruction is accurate. To account for this, we adopt a strategy similar to the test-time optimization used in NeRF-W~\cite{nerf_w} and WildGaussians~\cite{wGS}. With the trained GS model fully frozen, we search within $\pm\delta$ of $\tilde{p}_{test}^f$ for the phase $\tilde{p'}_{test}^f$ that yields the highest PSNR against the ground-truth frame. Because the search window is small and phase is a single scalar, this procedure corrects for estimator noise without introducing meaningful bias toward the ground truth. In the experiments, we set $\delta=0.05$, which corresponds to a window size of approximately 10 to 30 frames relative to the ground truth. The final metrics are computed from images rendered at these refined phases. To further assess the effect of synchronization on rendering quality, we compute metrics exclusively over dynamic regions. As shown in Fig.~\ref{fig:mask}, segmentation masks are extracted from the ground-truth frames, and PSNR, SSIM, and LPIPS are evaluated only within these masked areas. Since dynamic regions are most sensitive to temporal misalignment, restricting the evaluation to these areas allows for a more precise examination of whether the phase reference is accurately synchronized with the ground truth.

\paragraph{Search Window Sensitivity.}
To verify that the test-time phase search only compensates for the estimation noise, we vary the search window $W$ and report the PSNR with the FreeTimeGS~\cite{ftgs} backbone in Tab.~\ref{tab:search_window}. $W{=}0$ denotes rendering directly at the reference phase $\tilde{p}^{f}_{\text{test}}$ without any search, and $W{=}10$ is the default setting used for all reported results. Increasing the window from 10 to 50 does not change the PSNR in any setting, which indicates that the search converges within a small neighborhood of the reference phase. Without the search ($W{=}0$), the PSNR changes by less than 0.2\,dB in five of the six settings, and the largest drop of 0.75\,dB is observed in the Shifted condition of the Jumping scene. Even at $W{=}0$, Phase-DGS outperforms the FreeTimeGS baseline by 4.8--7.0\,dB under all Shifted and Dropped settings (see Tab.~\ref{Table_SelfCap} and~\ref{Table_Nvidia}), which shows that the performance gains do not result from the test-time search.

\begin{table}[h]
  \centering
  \caption{Sensitivity of PSNR to the search window $W$ of the test-time phase search with the FreeTimeGS backbone. $W{=}0$ denotes rendering at the reference phase without search, and $W{=}10$ is the default setting.}
  \label{tab:search_window}
  \begin{tabular}{llcccc}
    \toprule
    Condition & Scene & $W{=}0$ & $W{=}5$ & \makecell{$W{=}10$ \\ (default)} & $W{=}50$ \\
    \midrule
    \multirow{2}{*}{Synced}  & Bike    & 31.355 & 31.376 & 31.376 & 31.376 \\
                             & Jumping & 28.851 & 28.851 & 28.851 & 28.851 \\
    \midrule
    \multirow{2}{*}{Shifted} & Bike    & 31.110 & 31.299 & 31.299 & 31.299 \\
                             & Jumping & 27.354 & 28.090 & 28.103 & 28.103 \\
    \midrule
    \multirow{2}{*}{Dropped} & Bike    & 31.131 & 31.217 & 31.217 & 31.217 \\
                             & Jumping & 29.386 & 29.386 & 29.386 & 29.386 \\
    \bottomrule
  \end{tabular}
\end{table}

\begin{algorithm}[h]
\caption{Phase Reference Computation for Test Views}
\label{alg:phase_eval}
\begin{algorithmic}[1]
\Require Train view images $\{u_i^{f}\}$ with source indices $\{s_i(f)\}$, test view source indices $\{s_{\text{test}}(f)\}$, Trained Phase projector $\Phi$
\Ensure Reference phase $\tilde{p}_{\text{test}}^{f}$ for each test view frame
\For{each test view frame index $f = 1, \dots, F$}
    \State $\mathcal{T} \gets \emptyset$ \Comment{Collect matching train view frames}
    \For{each train view $i = 1, \dots, N$}
        \For{each frame index $f' = 1, \dots, F$}
            \If{$s_i(f') = s_{\text{test}}(f)$}
                \State $\mathcal{T} \gets \mathcal{T} \cup \{(i, f')\}$
            \EndIf
        \EndFor
    \EndFor
    \State $p_{\text{sum}} \gets 0$
    \For{each $(i, f') \in \mathcal{T}$}
        \State $p_{\text{sum}} \gets p_{\text{sum}} + \Phi(u_i^{f'})$ \Comment{Extract phase from each matching train frame}
    \EndFor
    \State $\tilde{p}_{\text{test}}^{f} \gets p_{\text{sum}} / |\mathcal{T}|$
\EndFor
\State \Return $\{\tilde{p}_{\text{test}}^{f}\}_{f=1}^{F}$
\end{algorithmic}
\end{algorithm}

\clearpage

\begin{figure}[!t]
    \centering
    \includegraphics[trim=7.8cm 7.2cm 7.8cm 2.5cm, clip, width=0.7\linewidth]{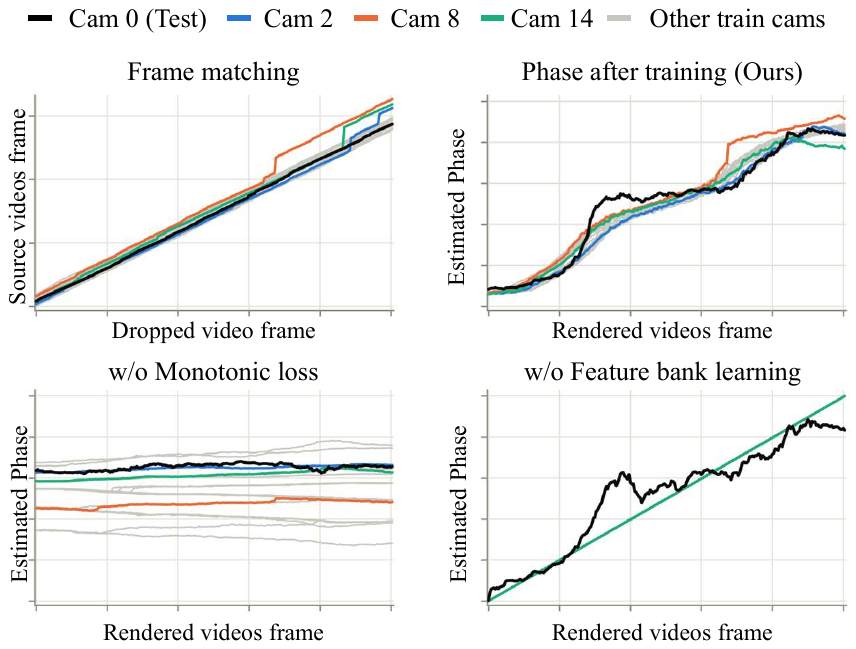}
    \caption{Per-frame phase estimates on the Bike scene (SelfCap) under the Dropped condition. (Top left) Ground-truth frame correspondences. Phase estimation results under (Top right) default setting, (Bottom left) without monotonic loss. (Bottom right) without feature bank learning.}
    \label{fig:Ablation Detail}
\end{figure}

\section{Additional Discussion}\label{sec:discussion}
\textbf{Backbone-Dependent Reconstruction Quality.} As briefly noted in Section~5, the performance of Phase-DGS is influenced by the visual representation capacity of the underlying 4DGS backbone. We attribute this to the relatively large weight assigned to the rendering loss during training, which makes the rendering quality of the backbone an important factor for stable phase learning. We provide a more detailed observation here. The RealTime-4DGS~\cite{4dgs} backbone began to overfit to the training scene after approximately 60,000 iterations, where the test-view PSNR dropped sharply and, at the same time, the consistency of the estimated phase values also deteriorated. In contrast, the FreeTimeGS~\cite{ftgs} backbone maintained stable visual quality up to 500,000 iterations, and the corresponding phase estimates remained close to the ground-truth phase profile throughout training.

\textbf{Global phase estimation in Practice.} The phase projector module of Phase-DGS estimates the phase independently for each frame of each camera view. As shown in our experiments, it is not possible to define an accurate global time stamp under the Dropped condition. Since Phase-DGS can only extract relative phase stamps, the global phase stamp must be defined externally. We propose two approaches for this. The first uses the average of the phase stamps estimated across all camera views as the global phase stamp. The second uses the phase estimated from a single reference camera view as the global phase stamp.

\section{Ablation Detail}

As shown in the ablation study, the learnable feature bank and Monotonic Loss are the two most critical components for training Phase-DGS. Removing either of them prevents accurate phase estimation, making the Gaussian Splatting training infeasible. Figure \ref{fig:Ablation Detail} shows the per-frame predicted phase of each camera view, produced by the fully trained phase projector module. In the default setting, the relative phase values across frames follow the ground truth labels. Without Monotonic Loss, phase differences across views appear, but the phase within a single view barely changes, making training collapse. Without feature bank learning, all train views produce the same linear phase output, which means the phase projector module provides no useful information.

\section{Numerical Stability}
To verify the stability of Phase-DGS across different runs, we train the model four times from scratch, including COLMAP initialization, on the \textit{Corgi} scene under the Dropped condition. As reported in Tab.~\ref{tab:stability}, Phase-DGS achieves consistent performance across all runs, with a PSNR standard 
deviation of 0.466 dB, demonstrating reliable convergence regardless of initialization.

\begin{table}[h]
\centering
\caption{Numerical stability of Phase-DGS across four independent runs on the 
\textit{Corgi} scene \textbf{(SelfCap dataset)} under the Dropped condition.}
\label{tab:stability}
\resizebox{0.4\linewidth}{!}{%
\begin{tabular}{lccc}
\toprule
Run & PSNR$\uparrow$ & SSIM$\uparrow$ & LPIPS$\downarrow$ \\
\midrule
1 & 35.81 & 0.989& 0.008\\
2 & 36.86 & 0.990& 0.007\\
3 & 36.04 & 0.990& 0.008\\
4 & 35.99 & 0.990& 0.008\\
\midrule
Mean  &   36.175 &   0.990 &   0.008 
\\
Std   &    0.466489 &   0.000450 &   0.000455 
\\
\bottomrule
\end{tabular}
}
\end{table}

\section{Failure Case}\label{sec:failure}
\begin{figure}[h]
    \centering
    \includegraphics[trim=0cm 2.5cm 0cm 3cm, clip, width=1\linewidth]{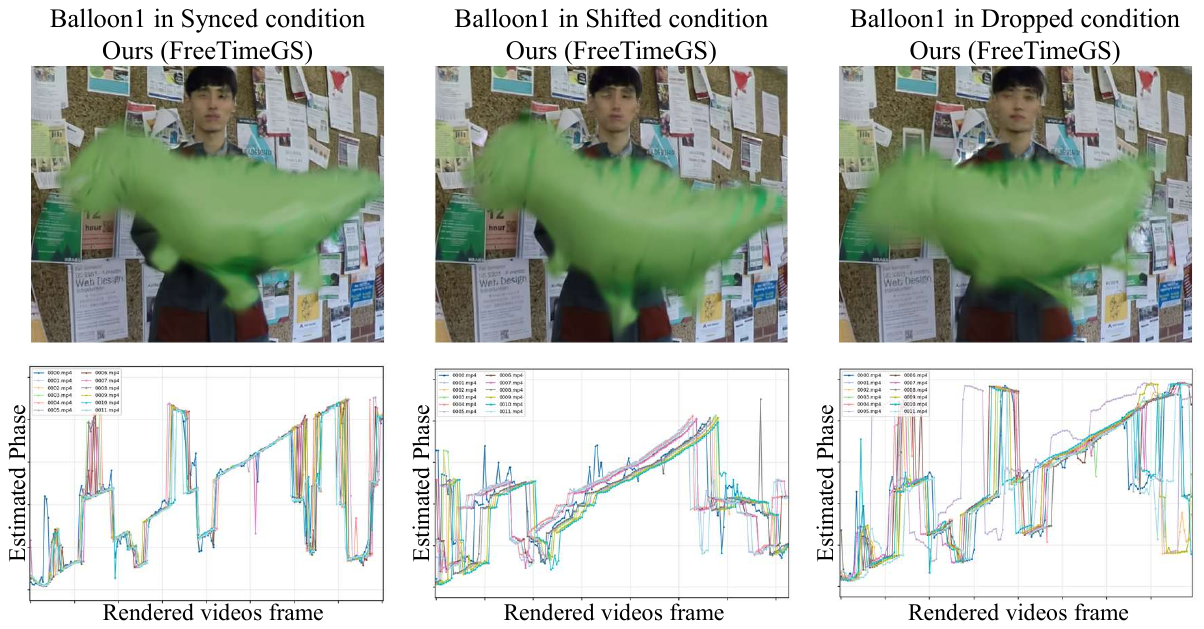}
    \caption{Failure Case of Phase-DGS on Balloon1, a Highly Periodic Scene. }
    \label{fig:failure}
\end{figure}

\textbf{Highly Periodic Content.} Since Phase-DGS estimates the phase from visual patterns extracted from image content, it has a limitation when a scene contains highly repetitive patterns along the temporal axis. We observed this behavior in the Balloon scene of the Nvidia Dynamic Scene Dataset~\cite{nvidia_dataset}. The scene consists of a person repeatedly moving a balloon up and down against a simple background, and the visual patterns of frames at different times remain nearly identical. As a result, the phase projector assigns similar phase values to frames that are temporally distant, which provides ambiguous supervision to the 4DGS training and prevents accurate reconstruction. Fig.~\ref{fig:failure} shows the estimated phase values and the rendered results on the Balloon scene. This limitation arises from the fact that Phase-DGS uses only the visual pattern of a single frame as input, and extending the phase estimation module to incorporate temporal causality across frames (e.g., frame indices or global frame-to-frame similarity) or cross-camera information is expected to mitigate this ambiguity.

\newpage 

\section{Per-scene quantitative results }
Tables~\ref{Table_SelfCap} and~\ref{Table_Nvidia} provide per-scene results corresponding to the averages reported in Tab.~\ref{Table_Avg}.
\input{Table_Final}

\newpage 

\section{License Information}
All datasets, pretrained models, and baseline implementations used in this work are used in accordance with their respective licenses and terms of use.

\textbf{Datasets.} We use the Nvidia Dynamic Scene Dataset~\cite{nvidia_dataset} and SelfCap~\cite{selfcap_dataset} for evaluation. Both datasets are publicly available for academic research and are used in accordance with the terms provided by their respective authors.

\textbf{Pretrained Models.} We use DINOv2~\cite{dinov2} as the ViT encoder, which is released under the Apache 2.0 license.

\textbf{Baseline Methods.} Our re-implementation of FreeTimeGS~\cite{ftgs} is based on the original paper. We also compare with RealTime-4DGS~\cite{4dgs}, VisualSync~\cite{visualsync}, and Sync-NeRF~\cite{syncnerf}. We cite the original papers for all methods and use them in accordance with their stated conditions.

\end{document}

%% file: Tables/Table_finalfinal.tex
\begin{table}[!t]
\centering
\caption{Average quantitative comparison on the \textbf{SelfCap} and \textbf{Nvidia Dynamic Scene} datasets under 
synchronized (Synced) and unsynchronized (Shifted, Dropped) conditions. 
$\uparrow$: higher is better, $\downarrow$: lower is better. 
\colorbox{green!20}{Green}: best among methods sharing the same backbone 
and condition. \textcolor{red}{\textbf{Red}}: best across all methods 
under the same condition.}
\label{Table_Avg}
\resizebox{0.9\columnwidth}{!}{%
\begin{tabular}{c c l ccc ccc}
\toprule
& & & \multicolumn{3}{c}{\textbf{SelfCap}}
  & \multicolumn{3}{c}{\textbf{Nvidia}} \\
\cmidrule(lr){4-6}\cmidrule(lr){7-9}
\textbf{Condition} & \textbf{Backbone} & \textbf{Method}
  & PSNR$\uparrow$ & SSIM$\uparrow$ & LPIPS$\downarrow$
  & PSNR$\uparrow$ & SSIM$\uparrow$ & LPIPS$\downarrow$ \\
\midrule
\multirow{6}{*}{Synced}
& \multirow{1}{*}{K-Planes}& Sync-NeRF  & - & - & - & 28.889 & 0.942 & 0.048 \\
\cmidrule(l){2-9}
& \multirow{2}{*}{RealTime-4DGS}& -  & 30.858 & 0.970 & 0.030 & 29.735\cellcolor{green!20} & 0.952\cellcolor{green!20} & 0.045\cellcolor{green!20} \\
& & Ours  & 32.145\cellcolor{green!20} & 0.973\cellcolor{green!20} & 0.026\cellcolor{green!20} & 28.938 & 0.946 & 0.061 \\
\cmidrule(l){2-9}
& \multirow{3}{*}{FreeTimeGS}& -  & 34.181 & 0.984
& \textcolor{red}{0.010}\cellcolor{green!20} & \textcolor{red}{31.167}\cellcolor{green!20} & \textcolor{red}{0.963}\cellcolor{green!20} & \textcolor{red}{0.022}\cellcolor{green!20} \\
& & VisualSync  & 33.446 & 0.983 & 0.013 & 30.429 & 0.958 & 0.025 \\
& & Ours  & \textcolor{red}{34.727}\cellcolor{green!20} & \textcolor{red}{0.985}\cellcolor{green!20} & 0.012 & 30.416 & 0.959 & 0.030 \\
\midrule
\multirow{6}{*}{Shifted}
& \multirow{1}{*}{K-Planes}& Sync-NeRF  & - & - & - & 28.492 & 0.940 & 0.049 \\
\cmidrule(l){2-9}
& \multirow{2}{*}{RealTime-4DGS}& -  & 27.614 & 0.956 & 0.039 & 23.083 & 0.915 & 0.084 \\
& & Ours  & 28.288\cellcolor{green!20} & 0.961\cellcolor{green!20} & 0.037\cellcolor{green!20} & 26.655\cellcolor{green!20} & 0.937\cellcolor{green!20} & 0.064\cellcolor{green!20} \\
\cmidrule(l){2-9}
& \multirow{3}{*}{FreeTimeGS}& -  & 27.170 & 0.955 & 0.026 & 23.785 & 0.918 & 0.068 \\
& & VisualSync  & 29.611 & 0.967 & 0.018 & 27.958 & 0.950 & \textcolor{red}{0.033}\cellcolor{green!20} \\
& & Ours  & \textcolor{red}{32.947}\cellcolor{green!20} & \textcolor{red}{0.981}\cellcolor{green!20} & \textcolor{red}{0.014}\cellcolor{green!20} & \textcolor{red}{29.836}\cellcolor{green!20} & \textcolor{red}{0.956}\cellcolor{green!20} & 0.037 \\
\midrule
\multirow{6}{*}{Dropped}
& \multirow{1}{*}{K-Planes}& Sync-NeRF  & - & - & - & 26.683 & 0.929 & 0.064 \\
\cmidrule(l){2-9}
& \multirow{2}{*}{RealTime-4DGS}& -  & 26.679 & 0.959 & 0.036 & 25.547 & 0.931 & 0.068 \\
& & Ours  & 29.638\cellcolor{green!20} & 0.966\cellcolor{green!20} & 0.032\cellcolor{green!20} & 28.263\cellcolor{green!20} & 0.944\cellcolor{green!20} & 0.056\cellcolor{green!20} \\
\cmidrule(l){2-9}
& \multirow{3}{*}{FreeTimeGS}& -  & 27.894 & 0.962 & 0.025 & 25.313 & 0.929 & 0.055 \\
& & VisualSync  & 29.739 & 0.972 & 0.019 & 28.057 & 0.943 & 0.039 \\
& & Ours  & \textcolor{red}{34.379}\cellcolor{green!20} & \textcolor{red}{0.984}\cellcolor{green!20} & \textcolor{red}{0.012}\cellcolor{green!20} & \textcolor{red}{30.665}\cellcolor{green!20} & \textcolor{red}{0.961}\cellcolor{green!20} & \textcolor{red}{0.030}\cellcolor{green!20} \\
\bottomrule
\end{tabular}%
}
\end{table}

%% file: Table_Ablation.tex
\begin{table}[h]
\centering
\caption{Ablation study on the \textit{Jumping} scene \
\textbf{(Nvidia Dynamic Scene dataset)} 
and \textit{Bike} scene \textbf{(SelfCap dataset)} under the Dropped condition. 
FreeTimeGS is used as the DGS backbone.}
\label{tab:ablation}
\resizebox{0.75\linewidth}{!}{%
\begin{tabular}{lcccccc}
\toprule
& \multicolumn{3}{c}{Jumping} & \multicolumn{3}{c}{Bike} \\
\cmidrule(lr){2-4} \cmidrule(lr){5-7}
Method & PSNR$\uparrow$ & SSIM$\uparrow$ & LPIPS$\downarrow$ 
       & PSNR$\uparrow$ & SSIM$\uparrow$ & LPIPS$\downarrow$ \\
\midrule
Ours (default)   & 29.386 & 0.942 & 0.037 & 31.212& 0.977 & 0.016 \\
\toprule
w/o Monotonic Loss   & 24.421& 0.887 & 0.073 & 25.903& 0.944 & 0.039 \\
w/o Bank Learning         & 24.794& 0.897 & 0.070 & 27.642& 0.958 & 0.026 \\
 $k=1$                & 29.425& 0.942 & 0.038 & 31.130& 0.977 &0.015 
\\
 $k=4$                & 29.006& 0.938 & 0.038 & 31.101& 0.976 &0.015 \\
DINOv2-S             & 29.034& 0.939 & 0.039 & 30.932& 0.976 & 0.016 
\\
DINOv2-L             & 29.648& 0.943 & 0.037 & 31.411& 0.978& 0.015\\
w/o ROI Crop         & 29.231& 0.939 & 0.039 & 31.16& 0.977 & 0.017\\
\bottomrule
\end{tabular}
}
\end{table}

%% file: Table_Final.tex
\begin{table*}[h]
\centering
\caption{Quantitative comparison on the \textbf{SelfCap dataset} under 
synchronized (Synced) and unsynchronized (Shifted, Dropped) conditions. 
$\uparrow$: higher is better, $\downarrow$: lower is better. 
\colorbox{green!20}{Green}: best among methods sharing the same backbone 
and condition. \textcolor{red}{\textbf{Red}}: best across all methods 
under the same condition.}
\label{Table_SelfCap}
\resizebox{\textwidth}{!}{%
\begin{tabular}{c c l ccc ccc ccc}
\toprule
& & & \multicolumn{3}{c}{\textbf{Bike}}
  & \multicolumn{3}{c}{\textbf{Corgi}}
  & \multicolumn{3}{c}{\textbf{Yoga}} \\
\cmidrule(lr){4-6}\cmidrule(lr){7-9}\cmidrule(lr){10-12}
\textbf{Condition}& \textbf{Backbone} & \textbf{Method}
  & PSNR$\uparrow$ & SSIM$\uparrow$ & LPIPS$\downarrow$
  & PSNR$\uparrow$ & SSIM$\uparrow$ & LPIPS$\downarrow$
  & PSNR$\uparrow$ & SSIM$\uparrow$ & LPIPS$\downarrow$ \\
\midrule

\multirow{7}{*}{Synced}
& \multirow{1}{*}{K-Planes}& Sync-NeRF  & - & - & - & - & - & - & - & - & - \\
\cmidrule(l){2-12}
& \multirow{2}{*}{RealTime-4DGS}
  & -& 26.430 & 0.950 & 0.040 & 35.552\cellcolor{green!20}& 0.985\cellcolor{green!20}& 0.013\cellcolor{green!20}& 30.593 & 0.974 & 0.036 \\
&
  & Ours        & 28.565\cellcolor{green!20}& 0.960\cellcolor{green!20}& 0.032\cellcolor{green!20}& 34.907 & 0.982 & 0.019 & 32.963\cellcolor{green!20}& 0.978\cellcolor{green!20}& 0.028\cellcolor{green!20}\\
\cmidrule(l){2-12}
& \multirow{3}{*}{FreeTimeGS}& -& 30.888 & 0.977 & \textcolor{red}{0.014} \cellcolor{green!20}& 35.677 & 0.990 & \textcolor{red}{0.006}\cellcolor{green!20}& \textcolor{red}{35.977} \cellcolor{green!20}& \textcolor{red}{0.987}\cellcolor{green!10}& \textcolor{red}{0.011}\cellcolor{green!20}\\
&
  & VisualSync  & 30.490 & 0.974 & 0.016 & 36.646 & 0.990 & \textcolor{red}{0.006}& 33.203 & 0.984 & 0.016 \\
&
  & Ours        & \textcolor{red}{31.376}\cellcolor{green!20}& \textcolor{red}{0.978}\cellcolor{green!20}& 0.015 & \textcolor{red}{37.681}\cellcolor{green!20}& \textcolor{red}{0.991}\cellcolor{green!20}& 0.007 & 35.124 & \textcolor{red}{0.987}\cellcolor{green!10}& 0.013 \\

\midrule

\multirow{7}{*}{Shifted}
& \multirow{1}{*}{K-Planes}& Sync-NeRF  & - & - & - & - & - & - & - & - & - \\
\cmidrule(l){2-12}
& \multirow{2}{*}{RealTime-4DGS}
  & -& 24.594 & 0.932 & 0.049 & 29.100 & 0.971 & 0.025 & 29.149\cellcolor{green!20}& 0.966\cellcolor{green!20}& 0.043\cellcolor{green!20}\\
&
  & Ours        & 26.753\cellcolor{green!20}& 0.948\cellcolor{green!20}& 0.036\cellcolor{green!20}& 30.812\cellcolor{green!20}& 0.974\cellcolor{green!20}& 0.021\cellcolor{green!20}& 27.298 & 0.961 & 0.054 \\
\cmidrule(l){2-12}
& \multirow{3}{*}{FreeTimeGS}& -& 24.139 & 0.935 & 0.035 & 27.678 & 0.965 & 0.018 & 29.693 & 0.965 & 0.026 \\
&
  & VisualSync  & 26.227 & 0.949 & 0.024 & 29.975 & 0.974 & 0.012 & 32.632 & 0.978 & 0.018 \\
&
  & Ours        & \textcolor{red}{31.299}\cellcolor{green!20}& \textcolor{red}{0.977}\cellcolor{green!20}& \textcolor{red}{0.016}\cellcolor{green!20}& \textcolor{red}{34.767}\cellcolor{green!20}& \textcolor{red}{0.988}\cellcolor{green!20}& \textcolor{red}{0.010}\cellcolor{green!20}& \textcolor{red}{32.775}\cellcolor{green!20}& \textcolor{red}{0.979}\cellcolor{green!20}& \textcolor{red}{0.017}\cellcolor{green!20}\\

\midrule

\multirow{7}{*}{Dropped}
& \multirow{1}{*}{K-Planes}& Sync-NeRF  & - & - & - & - & - & - & - & - & - \\
\cmidrule(l){2-12}
& \multirow{2}{*}{RealTime-4DGS}
  & -& 26.339 & 0.946 & 0.037 & 27.149 & 0.967 & 0.025 & 26.548 & 0.963\cellcolor{green!20}& 0.045\cellcolor{green!20}\\
&
  & Ours        & 28.608\cellcolor{green!20}& 0.961\cellcolor{green!20}& 0.029\cellcolor{green!20}& 33.664\cellcolor{green!20}& 0.981\cellcolor{green!20}& 0.016\cellcolor{green!20}& 26.642\cellcolor{green!20}& 0.956 & 0.052 \\
\cmidrule(l){2-12}
& \multirow{3}{*}{FreeTimeGS}& -& 26.304 & 0.947 & 0.027 & 26.505 & 0.964 & 0.020 & 30.872 & 0.976 & 0.028 \\
&
  & VisualSync  & 29.746 & 0.970 & 0.019 & 27.491 & 0.968 & 0.017 & 31.981 & 0.978 & 0.021 \\
&
  & Ours        & \textcolor{red}{31.217}\cellcolor{green!20}& \textcolor{red}{0.977}\cellcolor{green!20}& \textcolor{red}{0.016}\cellcolor{green!20}& \textcolor{red}{36.859}\cellcolor{green!20}& \textcolor{red}{0.990}\cellcolor{green!20}& \textcolor{red}{0.007}\cellcolor{green!20}
& \textcolor{red}{35.061}\cellcolor{green!20}& \textcolor{red}{0.985}\cellcolor{green!20}& \textcolor{red}{0.014}\cellcolor{green!20}\\

\bottomrule
\end{tabular}%
}
\centering
\caption{Quantitative comparison on the \textbf{Nvidia Dynamic Scene dataset} under 
synchronized (Synced) and unsynchronized (Shifted, Dropped) conditions. 
$\uparrow$: higher is better, $\downarrow$: lower is better. 
\colorbox{green!20}{Green}: best among methods sharing the same backbone 
and condition. \textcolor{red}{\textbf{Red}}: best across all methods 
under the same condition.}
\label{Table_Nvidia}
\resizebox{\textwidth}{!}{%
\begin{tabular}{c c l ccc ccc ccc}
\toprule
& & & \multicolumn{3}{c}{\textbf{Umbrella}}
  & \multicolumn{3}{c}{\textbf{Jumping}}
  & \multicolumn{3}{c}{\textbf{Playground}} \\
\cmidrule(lr){4-6}\cmidrule(lr){7-9}\cmidrule(lr){10-12}
\textbf{Condition}& \textbf{Backbone} & \textbf{Method}
  & PSNR$\uparrow$ & SSIM$\uparrow$ & LPIPS$\downarrow$
  & PSNR$\uparrow$ & SSIM$\uparrow$ & LPIPS$\downarrow$
  & PSNR$\uparrow$ & SSIM$\uparrow$ & LPIPS$\downarrow$ \\
\midrule

\multirow{7}{*}{Synced}
& \multirow{1}{*}{K-Planes}& Sync-NeRF& 31.467 & 0.968 & 0.028 & 24.228 & 0.889 & 0.088 & \textcolor{red}{30.972}& \textcolor{red}{0.969}& 0.027 \\
\cmidrule(l){2-12}
& \multirow{2}{*}{RealTime-4DGS}
  & -& 32.494\cellcolor{green!20}& 0.971\cellcolor{green!20}& 0.026\cellcolor{green!20}& 26.891\cellcolor{green!20}& 0.917\cellcolor{green!20}& 0.073\cellcolor{green!20}& 29.820\cellcolor{green!20}& 0.967\cellcolor{green!20}& 0.035\cellcolor{green!20}\\
&
  & Ours& 30.328 & 0.961 & 0.049 & 26.724 & 0.912 & 0.093 & 29.763 & 0.965 & 0.040 \\
\cmidrule(l){2-12}
& \multirow{3}{*}{FreeTimeGS}& -& 33.839& \textcolor{red}{0.978}\cellcolor{green!20}& \textcolor{red}{0.019}\cellcolor{green!20}& \textcolor{red}{29.759}\cellcolor{green!20}& \textcolor{red}{0.942}\cellcolor{green!20}& \textcolor{red}{0.032}\cellcolor{green!20}& 29.904\cellcolor{green!20}& \textcolor{red}{0.969}\cellcolor{green!20}& \textcolor{red}{0.016}\cellcolor{green!20}\\
&
  & VisualSync& \textcolor{red}{33.998}\cellcolor{green!20}& \textcolor{red}{0.978}\cellcolor{green!20}& \textcolor{red}{0.019}\cellcolor{green!20}& 27.904& 0.929 & 0.039 & 29.384 & 0.967 & 0.018\\
&
  & Ours& 32.549 & 0.972 & 0.032 & 28.851 & 0.936 & 0.038 & 29.849 & \textcolor{red}{0.969}\cellcolor{green!20}& 0.021 \\

\midrule

\multirow{7}{*}{Shifted}
& \multirow{1}{*}{K-Planes}& Sync-NeRF& 30.947 & 0.967 & \textcolor{red}{0.030}& 25.088 & 0.891 & 0.087 & 29.442 & 0.963 & 0.031 \\
\cmidrule(l){2-12}
& \multirow{2}{*}{RealTime-4DGS}
  & -& 24.432 & 0.942 & 0.067 & 19.365 & 0.852 & 0.135 & 25.452 & 0.950 & 0.050 \\
&
  & Ours        & 29.215\cellcolor{green!20}& 0.958\cellcolor{green!20}& 0.047\cellcolor{green!20}& 25.124\cellcolor{green!20}& 0.901\cellcolor{green!20}& 0.097\cellcolor{green!20}& 25.627\cellcolor{green!20}& 0.951\cellcolor{green!20}& 0.047\cellcolor{green!20}\\
\cmidrule(l){2-12}
& \multirow{3}{*}{FreeTimeGS}& -& 24.169 & 0.945 & 0.068 & 20.985 & 0.860 & 0.100 & 26.202 & 0.950 & 0.036 \\
&
  & VisualSync  & 26.042 & 0.951 & 0.044 & 28.071 & \textcolor{red}{0.931}\cellcolor{green!20}& \textcolor{red}{0.038}& \textcolor{red}{29.761}& \textcolor{red}{0.967}& \textcolor{red}{0.018}\\
&
  & Ours        & \textcolor{red}{32.316}\cellcolor{green!20}& \textcolor{red}{0.971}\cellcolor{green!20}& 0.035\cellcolor{green!20}& \textcolor{red}{28.103}\cellcolor{green!20}& \textcolor{red}{0.931}\cellcolor{green!20}& 0.049\cellcolor{green!20}& 29.090\cellcolor{green!20}& 0.965\cellcolor{green!20}& 0.026\cellcolor{green!20}\\

\midrule

\multirow{7}{*}{Dropped}
& \multirow{1}{*}{K-Planes}& Sync-NeRF  & 29.708 & 0.955 & 0.053 & 22.175 & 0.873 & 0.104 & 28.167 & 0.960 & 0.035 \\
\cmidrule(l){2-12}
& \multirow{2}{*}{RealTime-4DGS}
  & -& 23.990 & 0.942 & 0.067 & 23.901 & 0.888 & 0.100 & 28.750& 0.964\cellcolor{green!20}& 0.038\cellcolor{green!20}\\
&
  & Ours        & 29.410\cellcolor{green!20}& 0.958\cellcolor{green!20}& 0.047\cellcolor{green!20}& 26.551\cellcolor{green!20}& 0.913\cellcolor{green!20}& 0.081\cellcolor{green!20}& 28.827\cellcolor{green!20}& 0.962 & 0.039 \\
\cmidrule(l){2-12}
& \multirow{3}{*}{FreeTimeGS}& -& 23.948 & 0.942 & 0.065 & 23.509 & 0.881 & 0.076 & 28.482 & 0.963 & 0.023 \\
&
  & VisualSync  & 30.069 & 0.965 & 0.036 & 25.605 & 0.903 & 0.058 & 28.497 & 0.962 & 0.023 \\
&
  & Ours        & \textcolor{red}{32.357}\cellcolor{green!20}& \textcolor{red}{0.970}\cellcolor{green!20}& \textcolor{red}{0.033}\cellcolor{green!20}& \textcolor{red}{29.386}\cellcolor{green!20}& \textcolor{red}{0.942}\cellcolor{green!20}& \textcolor{red}{0.037}\cellcolor{green!20}& \textcolor{red}{30.251}\cellcolor{green!20}& \textcolor{red}{0.971}\cellcolor{green!20}& \textcolor{red}{0.021}\cellcolor{green!20}\\

\bottomrule
\end{tabular}%
}
\end{table*}